\documentclass[table]{article}

\PassOptionsToPackage{numbers, compress}{natbib}
\usepackage[preprint]{neurips_2025}

\usepackage[utf8]{inputenc} % allow utf-8 input
\usepackage[T1]{fontenc}    % use 8-bit T1 fonts
\usepackage{hyperref}       % hyperlinks
\usepackage{url}            % simple URL typesetting
\usepackage{booktabs}       % professional-quality tables
\usepackage{amsfonts}       % blackboard math symbols
\usepackage{nicefrac}       % compact symbols for 1/2, etc.
\usepackage{microtype}      % microtypography
\usepackage{graphicx}
\usepackage{natbib}
\usepackage{latexsym}
\usepackage{amssymb}
\usepackage{amsmath}
\usepackage{amsthm}
\usepackage{booktabs}
\usepackage{enumitem}
\usepackage{graphicx}
\usepackage{color}
\usepackage{adjustbox}
\usepackage{multirow}
\usepackage{algorithm}
\usepackage{algorithmic}
\usepackage{amsfonts}
\usepackage{booktabs}
\usepackage[table]{xcolor}
\usepackage{placeins}

\title{FlexEdit: Marrying Free-Shape Masks to VLLM for Flexible Image Editing}

\begin{document}

% \author{%
%   Tianshuo Yuan\footnotemark[1] \\
%   Shenzhen Institute of Advanced Technology, CAS \\
%   \texttt{tianshuoy@outlook.com} \\
%   \And
%   Yuxiang Lin\thanks{Equal contribution.} \\
%   Shenzhen Technology University \\
%   Georgia Institute of Technology \\
%   \texttt{yuxiang.lin@gatech.edu} \\
%   \And
%   Jue Wang \\
%   Shenzhen Institute of Advanced Technology, CAS \\
%   Shenzhen Technology University \\
%   \texttt{j.wang2@siat.ac.cn} \\
%   \And
%   Zhiqi Cheng \\
%   Carnegie Mellon University \\
%   \texttt{zhiqics@uw.edu} \\
%   \AND
%   Xiaolong Wang \\
%   Shenzhen Technology University \\
%   \texttt{wangxiaolong2023@email.szu.edu.cn} \\
%   \And
%   Guohua Jiao \\
%   Shenzhen Institute of Advanced Technology, CAS \\
%   \texttt{gh.jiao@siat.ac.cn} \\
%   \And
%   Wei Chen \\
%   Shenzhen Institute of Advanced Technology, CAS \\
%   \texttt{chenwei@siat.ac.cn} \\
%   \And
%   Xiaojiang Peng\thanks{Corresponding author.} \\
%   Shenzhen Technology University \\
%   \texttt{Pengxiaojiang@sztu.edu.cn} \\
% }

\author{%
  Tianshuo Yuan\textsuperscript{1}\footnotemark[1] \\
  \texttt{tianshuoy@outlook.com} \\
  \And
  Yuxiang Lin\textsuperscript{2,3}\thanks{These authors contributed equally.} \\
  \texttt{yuxiang.lin@gatech.edu} \\
  \And
  Jue Wang\textsuperscript{1,2} \\
  \texttt{j.wang2@siat.ac.cn} \\
  \And
  Zhiqi Cheng\textsuperscript{4} \\
  \texttt{zhiqics@uw.edu} \\
  \And
  Xiaolong Wang\textsuperscript{2} \\
  \texttt{wangxiaolong2023@email.szu.edu.cn} \\
  \And
  Guohua Jiao\textsuperscript{1} \\
  \texttt{gh.jiao@siat.ac.cn} \\
  \And
  Wei Chen\textsuperscript{1} \\
  \texttt{chenwei@siat.ac.cn} \\
  \And
  Xiaojiang Peng\textsuperscript{2}\thanks{Corresponding author.} \\
  \texttt{Pengxiaojiang@sztu.edu.cn} \\
}

\maketitle

\begin{center}
    \textsuperscript{1}Shenzhen Institute of Advanced Technology, Chinese Academy of Sciences \quad \\
    \textsuperscript{2}Shenzhen Technology University \\
    \textsuperscript{3}Georgia Institute of Technology \quad \\
    \textsuperscript{4}Carnegie Mellon University
\end{center}

\begin{abstract}
% 将视觉大型语言模型（VLLM）与扩散模型相结合，为根据人类语言指令执行复杂的图像编辑任务提供了一种强大的方法。然而，仅靠语言指令往往无法准确传达用户需求，尤其是当用户想要添加、替换或删除图像特定区域的元素时。遮罩等视觉提示可以有效、清晰地指出要编辑的具体位置或元素，从而克服语言指令的局限性。目前，大多数遮罩参考方法都要求用户在所需位置精确绘制图形，这对用户非常不友好。为了解决这个问题，我们提出了 IdealEdit，这是一种端到端的图像编辑方法，既能利用自由形状掩码进行视觉提示，又能利用语言指令。我们的方法采用 VLLM 来理解图像内容、视觉提示和用户指令。此外，我们还引入了一种适配器结构，将 VLLM 的嵌入与图像数据融合在一起，确保视觉信息与模型输出的无缝集成。此外，我们还构建了一个新的评估数据集，专门用于视觉提示和复杂指令。广泛的实验表明，我们的方法在基于 LLM 的图像编辑方面达到了最先进（SOTA）的性能，而且我们的简单提示技术也非常有效。
%代码和数据不久后将会开源.
Combining Vision Large Language Models (VLLMs) with diffusion models offers a powerful method for executing image editing tasks based on human language instructions. However, language instructions alone often fall short in accurately conveying user requirements, particularly when users want to add, replace elements in specific areas of an image. Luckily, masks can effectively indicate the exact locations or elements to be edited, while they require users to precisely draw the shapes at the desired locations, which is highly user-unfriendly. To address this, we propose FlexEdit, an end-to-end image editing method that leverages both free-shape masks and language instructions for \textbf{Flex}ible \textbf{Edit}ing. Our approach employs a VLLM in comprehending the image content, mask, and user instructions. Additionally, we introduce the Mask Enhance Adapter (MEA) that fuses the embeddings of the VLLM with the image data, ensuring a seamless integration of mask information and model output embeddings. Furthermore, we construct FSMI-Edit, a benchmark specifically tailored for free-shape mask, including 8 types of free-shape mask. Extensive experiments show that our method achieves state-of-the-art (SOTA) performance in LLM-based image editing, and our simple prompting technique stands out in its effectiveness.
The code and data can be found at \url{https://anonymous.4open.science/r/FlexEdit}
\end{abstract}
\newpage

%%%%%%%%%%%%%%%%%%%%%%%%%%%%%%%%%%%%%%%%%%%%%%%%%%%%%%%%%%%%%%%%%%%%%%%%
\section{Introduction}
% Image Editing工作得益于Diffusion Model的出色生成能力【引用】；通过在大量的text-image paries 训练后，这些模型可以生成与text prompts相吻合的高质量图片。Instruction-based 图像编辑将text prompt进一步扩展到了language instructions，以人类指令为依据对原图中的特定要素进行编辑~\cite{}。Despite these advancements, researchers have found that human instructions in image editing can often be ambiguous and complex.
\label{introduction}
% 1. 位置准
% 2. prompt简单

Image editing has significantly benefited from the exceptional generative capabilities of diffusion models~\cite{any2pix, anydoor, controlnet_inexplicit_mask, diffusion_models_beat_gans, gres, high_resdiffusion, instructdiffusion, instructpix2pix, ip-adapter, largen, magicbrush}. Trained on large scale datasets of text-image pairs, these models can generate high-quality images that align with text prompts. Instruction-based image editing extends these capabilities by using human language instructions to modify specific elements within an original image. However, researchers have found that language instructions can often be ambiguous and complex~\cite{lisa, pi2023detgpt, smartedit}, thus the text encoders~\cite{clip} fail to accurately represent this information, resulting in subpar editing outcomes. Large Language Models (LLMs)~\cite{gpt3, instructGPT}, with their strong instruction-following capabilities, can understand complex human instructions and reason out clear and simple editing directives. Models like SmartEdit~\cite{smartedit} have pioneered the use of LLMs as text encoders to comprehend complex instructions, allowing users to input complex commands for image editing. 
\begin{figure}[h]
    \centering
    \includegraphics[width=0.8\linewidth]{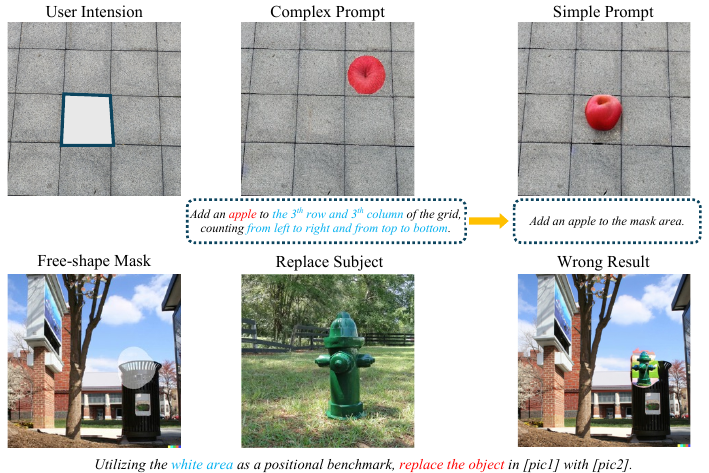}
    \caption{\small Free-shape mask can simplify the user input when dealing with location related instruction, and more flexible for user input. The compared methods are SmartEdit and Largen respectively.}
    \label{precise}
\end{figure}

\begin{figure*}[h]
    \centering
    \includegraphics[width=1.0\linewidth]{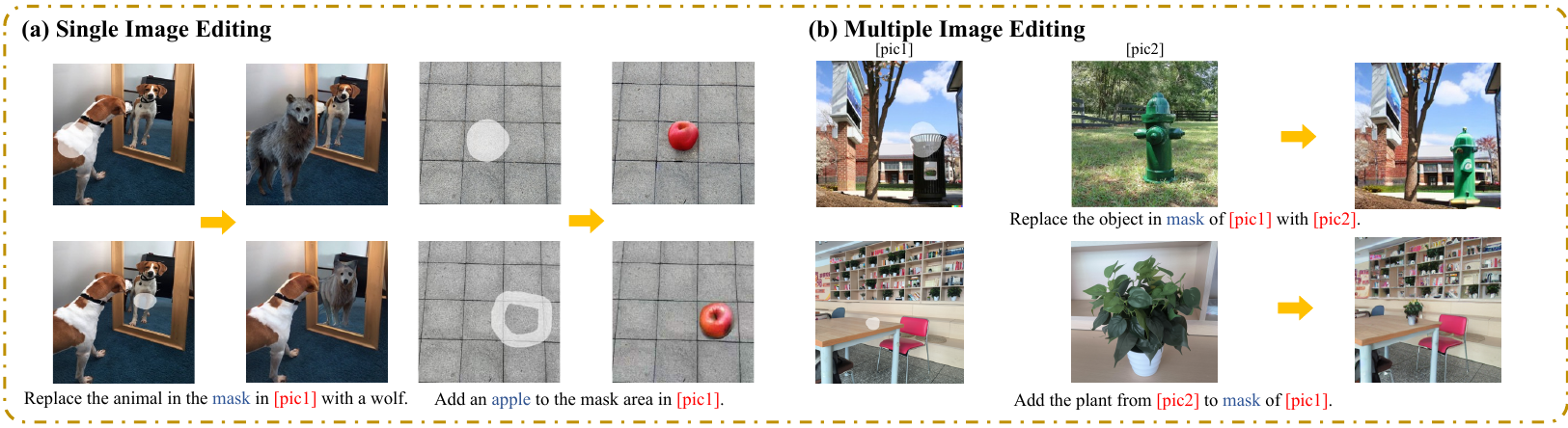}
    \caption{\small FlexEdit support both (a) single image and (b) multiple image editing, including replace/add an object. The free-shape mask input indicate the localization information, highly user-friendly than asking user to input full mask.}
    \label{idealedit}
\end{figure*}
However, when it comes to scenarios that are difficult to describe with language, such as specifying precise locations within an image, user have no idea how to prompt the models. Even if the model can understand, such prompts are clearly not user-friendly (e.g., ``$3^{th}$ row and $3^{th}$ column of the grid''). Users spend excessive time crafting these prompts, and the model may still fail to grasp their meaning, leading to unsatisfactory results (see Upper Figure~\ref{precise}). 
To address these challenges, visual prompts, such as masks, can effectively and clearly indicate the exact locations or elements to be edited, overcoming the limitations of language commands. 
% 修改开始：增加了HCI背景和引文，以加强动机。
The challenge of reducing the "tedious" and "time-consuming" effort of manual segmentation is well-documented in HCI literature, with foundational works like Intelligent Scissors~\cite{mortensen1995intelligent} and GrabCut~\cite{rother2004grabcut} created to simplify the process. Our work continues this user-centric philosophy. However, most contemporary methods that utilize mask references still require users to precisely draw shapes at the desired locations, which is highly user-unfriendly and can be a barrier to effective use~\cite{anydoor}. 
% 中文翻译如下：
% 在人机交互（HCI）的文献中，如何减少手动分割“繁琐”和“耗时”的工作这一挑战早有记载，例如“智能剪刀”~\cite{mortensen1995intelligent} 和 “GrabCut”~\cite{rother2004grabcut} 等开创性工作就是为了简化这一过程而创建的。我们的工作延续了这种以用户为中心的理念。然而，大多数当代使用掩模参考的方法仍然要求用户在期望位置精确绘制形状，这对用户非常不友好，并可能成为有效使用的障碍~\cite{anydoor}。
% 修改结束
If the mask shapes are irregular or do not completely overlap with the objects, they often produce very poor results (see Lower Figure~\ref{precise}). 

In real-world scenarios, the shapes users draw for masks are often random and may cover only a small part of the intended area. These free-shape masks challenge the model's ability to understand user requirements. To address this issue, we created the \textbf{F}ree \textbf{S}hape \textbf{M}ask \textbf{I}nstruction \textbf{Edit} (FSIM-Edit) benchmark, which includes 80 and 125 single images and multiple images respectively, 8 types of free-shape mask. Evaluating existing models with this benchmark revealed that they perform poorly in free-shape mask setting, even after the fine-tuning. 
To overcome these issues, we modified the mask functionality in MagicBrush~\cite{magicbrush}, integrating complex instructions with image editingin the FSIM-Edit dataset and proposed FlexEdit, an innovative end-to-end image editing method that leverages both free-shape masks and language instructions for \textbf{Flex}ible \textbf{Edit}ing (FlexEdit). FlexEdit employs a Vision Large Language Model (VLLM) to comprehend the image content, visual prompts, and user instructions. We introduced an adapter structure that fuses the mask represented embeddings of the VLLM with the image data, ensuring a seamless integration of visual information and model output embeddings. This integration enhances the model's ability to understand and execute complex editing tasks accurately. 
Experiments demonstrate that FlexEdit significantly outperforms other methods under free-shape mask conditions. Notably, our simple prompting technique proves to be highly effective, setting a new standard for ease of use and precision in the field. FlexEdit represents a significant step forward in the realm of intelligent image editing, combining the strengths of VLLMs and diffusion models to provide a user-friendly and highly capable editing solution. 
Our contributions can be summarized as follows:
\begin{itemize}
    \item We propose FlexEdit, an end-to-end image editing method that combines free-shape masks and language instructions, overcoming the limitations of traditional methods that require precise mask drawing. 
    \item We introduce the Mask Enhanced Adapter (MEA) structure that seamlessly enhance the mask embedding of the Vision Large Language Model (VLLM) with image data, enhancing the model's ability to understand and execute complex editing tasks. 
    \item We create the Free Shape Mask Instruction Edit (FSIM-Edit) benchmark, which includes a comprehensive dataset with diverse scenarios and editing instructions, to rigorously evaluate the performance of image editing models under free-shape mask conditions. 
\end{itemize}
\section{Related Works}
\label{sec:related_work}
In this section, we review the related works in image editing using diffusion models and the advancements in powerful Vision Large Language Models in image editing. 
\subsection{Image Editing with Diffusion Models}
% 修改开始：增加了关于 GANs 与扩散模型的讨论。
While our work is situated within the paradigm of VLLM-guided diffusion models, it is worth noting the contributions of Generative Adversarial Networks (GANs) to image editing. GAN-based methods often offer faster inference speeds. However, diffusion models generally provide superior training stability, mode coverage, and higher generation fidelity, which are critical for the complex, instruction-guided editing tasks that we target. Therefore, we focus on diffusion models and their direct state-of-the-art competitors for the most relevant and fair comparison.
% 中文翻译如下：
% 尽管我们的工作属于VLLM引导的扩散模型范式，但值得一提的是生成对抗网络（GANs）在图像编辑领域的贡献。基于GAN的方法通常提供更快的推理速度。然而，扩散模型通常在训练稳定性、模式覆盖率和生成保真度方面表现更优，而这些对于我们所针对的复杂、指令引导的编辑任务至关重要。因此，为了进行最相关和公平的比较，我们专注于扩散模型及其直接的SOTA竞争者。
% 修改结束

Image editing has achieved exceptional performance thanks to the generative abilities of diffusion models~\cite{instructpix2pix, magicbrush, instructdiffusion, anydoor, largen, ip-adapter}. \cite{instructpix2pix} introduced InstructPix2Pix, an instruction-based image editing approach that modifies images based on human language instructions. To better align with human instructions, \cite{magicbrush} proposed MagicBrush, extending text-only instructions to scenarios where a visual mask is provided. 
On the other hand, subject-driven image editing tasks aim to generate images conditioned on a customized subject and a text prompt that describes the context. For example, Anydoor~\cite{anydoor} can teleport target objects to new scenes at user-specified locations (mask) with desired shapes. \cite{largen} proposed Largen for image inpainting that enables seamless inpainting of masked scene images, incorporating both the textual prompts and specified subjects. 
\cite{smartmask} and \cite{controlnet_inexplicit_mask} start discussing to use coarse visual prompt such as bounding boxes, scribble in guiding the editing, but they use the classic text encoder in diffusion models, which makes it hard to understand those complex instructions. These models can combine user-provided images with visual prompts, such as masks, to perform image editing. The masks they use need to precisely overlap with the area to be edited, which is not user-friendly. 
In comparison, we propose using a free-shape mask as a visual prompt and introduce vision large language models (VLLMs) to better understand complex scenarios and instructions for visual prompt-based image editing in an end to end manner. 
\subsection{VLLMs in Image Editing}
Vision Large Language Models (VLLMs) are known for their extensive world knowledge and ability to understand complex instructions~\cite{qwen, vllava, llava, llava15, li2023otter, minigpt4v2}. Fine-tuning these VLLMs for image generation tasks has shown great success in comprehending human language instructions and generating images~\cite{smartedit, seedx, any2pix}. For example, \cite{smartedit} fine-tune the LLaVA~\cite{llava} model with dataset of MagicBrush~\cite{magicbrush} to fit it into the image editing domain. While InstructAny2Pix~\cite{any2pix} build a multi-modal editing system that enables users to edit an image using instructions involving audio, images, and text. 
Despite their success, current models predominantly rely on language-based instructions, overlooking the power of visual prompts such as masks. This reliance can limit the intuitiveness and accessibility of the editing process. Our approach seeks to enhance the editing capabilities by incorporating both user instructions and visual prompts, thereby achieving more precise and user-friendly image edits.
% \begin{figure*}[t]
%     \centering
%     \includegraphics[width=1.0\linewidth]{images/model.pdf}
%     \caption{Caption}
%     \label{model}
% \end{figure*}

%zhiqic
\begin{figure*}[t]
    \centering
    \includegraphics[width=1.0\linewidth]{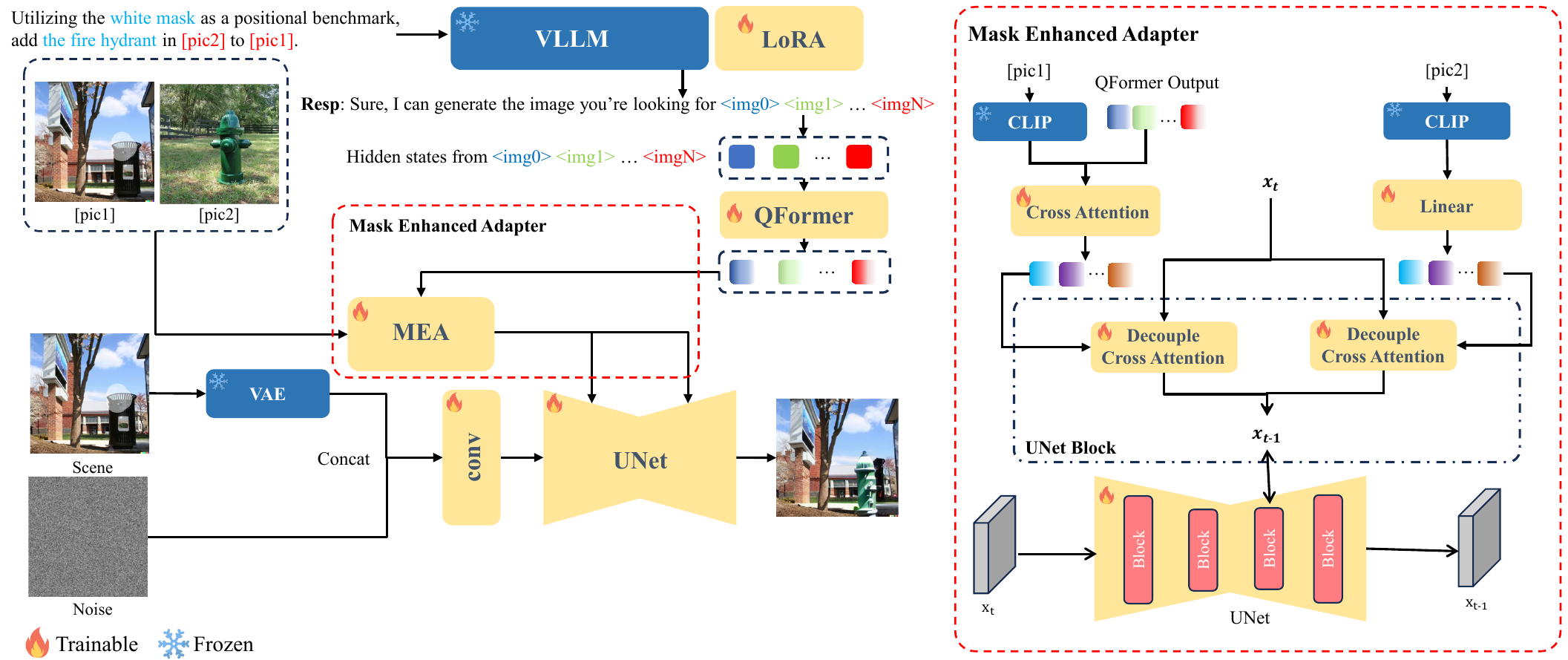}
    \caption{\small The architecture of the FlexEdit framework. FlexEdit integrates visual prompts and human instructions for complex image editing. It utilizes a VLLM backbone for multi-modal instruction understanding, a Q-Former for refining the hidden states, and a Mask Enhanced Adapter (MEA) for merging image and language model outputs. The final image generation is achieved through a diffusion model.}
    \label{fig:model}
\end{figure*}

% \section{FlexEdit: More Flexible Editing}
\section{Methodology}

This paper presents \textit{FlexEdit} designed to integrate human instructions and visual prompts for complex image editing tasks. Existing methods, whether based on language instructions or subject mask editing, struggle with understanding user-provided free-shape masks~\cite{anydoor,smartedit}. This challenge lead to a situation that user struggle to draw a mask to meet the models' need, which is not user-friendly and not acceptable. 

%zhiqic
In this section, We begin by introducing the FlexEdit Framework and explaining how it integrates free-shape mask with human instructions. We then discuss the Mask Enhanced Adapter (MEA) Module, which plays a crucial role in fusing image features. Finally, we explain the training process for understanding free-shape masks and present the FSMI-Edit Benchmark for evaluating our method.

\subsection{FlexEdit Framework}
\label{sec:flexedit_framework}
 The framework processes three inputs: a scene image $x_1$ with an associated mask $m$, a subject image $x_2$, and an editing instruction $\tau$. The primary objective is to generate a target image that incorporates $x_2$ and $\tau$ as the content, while utilizing $m$ as a positional reference within $x_1$. Importantly, the mask $m$ can take on a free-form shape and does not need to fully overlap with the intended position.
% Given a scene image $x_1$ with mask $m$, a subject image $x_2$, and an instruction $\tau$, our goal is to generate a target image using $x_2$ and $\tau$ as content, with $x_1$'s mask as a positional reference. The mask does not need to fully overlap with the intended position but can be a free-shape mask. 

We expanded the vocabulary of LLaVA~\cite{llava} with $N$ new tokens ``\textless$img0$\textgreater, \ldots, \textless$imgN$\textgreater'' to represent image editing information. As shown in Figure~\ref{fig:model}, $x_1$, $x_2$, $m$, and $\tau$ are input into the Vision Large Language Model (VLLM) $V(\cdot; \omega)$ to obtain response tokens $r_i$. 
Following this, the response set $R = \{r_1, r_2, \dots, r_n\}$ is processed to extract the corresponding hidden states $h_i \in \mathcal{H}$, capturing the deeper semantic information from the VLLM's understanding of the inputs. However, these hidden states primarily reside in the LLM's text vector space, which poses compatibility challenges when interfacing with the diffusion model-particularly one trained on CLIP~\cite{clip} text embeddings. To address this, we introduce a Q-Former~\cite{blip2} module that refines the hidden states into embeddings $e$ compatible with the diffusion model. The transformation process is summarized as follows:
% The response set $R=\{r_1, r_2, \dots, r_n\}$ is extracted, and $N$ image tokens $R^t = \{r_{N-j}, \dots, r_n\}$ are selected. Their corresponding hidden states $h_i \in \mathcal{H}$ are obtained using function $H$, where $\mathcal{H}$ represents all the corresponding hidden state matrices. These hidden states $\mathcal{H}$ are primarily in the LLM text vector space, making them difficult for the diffusion model (trained with CLIP~\cite{clip} text embeddings) to understand. To address this, we use an Q-Former~\cite{blip2} $Q$ to refine $\mathcal{H}$ to the embedding $e$, making them compatible with the diffusion model. This process is represented as:

\begin{equation}
    \begin{aligned}
    R~ &= V(x_1, x_2, \tau, m; \omega) \\
    R^t &= \{\text{``\textless$img{0 \dots N}$\textgreater''}\} \in R \\
    h_i &= H(x_1, x_2, \tau, m; \omega|R^t_i) \\
    e~ &= Q(\mathcal{H})
    \end{aligned}
    \label{image_embeddings}
\end{equation}
where $R$ is the response set, $R^t$ is the set of image tokens, and $e$ represents the embeddings transformed by the Q-Former function $Q$.
% $V(\cdot; \omega)$ represents the VLLM function with parameters $\omega$. The response tokens are $r_i$, and the response set is $R$. $R^t$ is the set of image tokens, and $h_i$ are the hidden states, with $\mathcal{H}$ representing the hidden state matrices. $Q$ is the Q-Former function, and $e$ are the CLIP text encoder embeddings.

For efficient training, we utilize LoRA~\cite{lora} fine-tuning techniques, freezing most VLLM parameters. Let the ground truth text label of the VLLM response be $R'$. The VLLM optimization process follows the recurrent loss calculation:
\begin{equation}
\mathcal{L}_{\text{VLLM}}(x, \tau) = -\sum_{\{(x, \tau), R'\}}\log p_{\omega+\Delta \omega(\theta)}(R' \mid (x, \tau))
\end{equation}
where $\theta$ represents the LoRA parameters.

Subsequently the Mask Enhanced Adapter (MEA) $A(\cdot; \delta)$ takes embedding $e$ and images $x_1$, $x_2$ as input, enhancing the mask feature from the VLLM with scene and subject images. The output $c$ from the MEA UNet block serves as the key and value for cross-attention in the subsequent block recurrently, calculated as:
\begin{equation}
c = A(x_1, x_2, e; \delta)
\end{equation}
here $A(\cdot; \delta)$ is the MEA function with parameters $\delta$, and $c$ is the MEA UNet block's output.

During the diffusion process, the encoded image latent $z = \mathcal{E}(x_1)$ is concatenated with the noisy latent $z_t$ and fed into the UNet~\cite{unet} in a residual manner. The UNet $\epsilon_{\alpha}$ is trained to predict the noise added to the noisy latent $z_t$, with noise levels increasing under timesteps $t$, given the condition $c$ from the MEA module. The diffusion optimization process is formulated as:
\begin{equation}
    \begin{split}
\mathcal{L}_{\text{diffusion}} &= \mathbb{E}_{\mathcal{E}(y), \mathcal{E}(x_1), c_T, \epsilon \sim \mathcal{N}(0, 1), t} \Big[ \| \epsilon - \\
&\epsilon_\alpha ( t, \text{concat}[z_t,  \mathcal{E}(x_1)], c ) \|_2^2 \Big]
    \end{split}
\end{equation}
where $\epsilon$ is the unscaled noise, $t$ is the sampling step, $z_t$ is the latent noise at step $t$, $\mathcal{E}(x_1)$ is the encoded scene image, and $c$ is the output from the MEA module.

The overall loss function combines VLLM and diffusion losses:
\begin{equation}
    \mathcal{L}_{\text{overall}} = \mathcal{L}_{\text{diffusion}} + \mathcal{L}_{\text{VLLM}}
\end{equation}

\subsection{Mask Enhanced Adapter}
\label{sec:mea}
% 好像缺了motivation

%场景和子图片首先会通过image encoder来映射到text space.接下来，the output from Qformer interacts with the feature from scene image. the 

The Mask Enhanced Adapter (MEA) Module aims at fusing the mask editing information from VLLM with the scene and subject images, enhancing the feature for fully interaction. Passing them to the diffusion models allows it to generate the result image for users.

% 补一下符号, decouple公式
The scene $x_1$ and subject $x_2$ images are first mapped to the text embedding space via the CLIP image encoder, for CLIP's training objective is to maximize the similarity of the image and text label embedding. The output $e$ of QFormer contains the editing intent generated by VLLM after combining the mask and instruction. However, it lacks fine-grained details from the original image. To reintroduce this information, we employed a cross-attention mechanism to fuse the output $e$ from QFormer~\cite{blip2} with the features from the scene image $x_1$, resulting the scene features $f_1$.
%因为qfomer的输出包含了VLLM的编辑意图用以引导图像编辑，所以
The key and value of the cross-attention come from the scene image $x_1$ with the the output $e$ from QFormer as the query. Formulated as follow:
\begin{equation}
    f_1 = \text{Softmax}\left(\frac{Q K^\top}{\sqrt{d}}\right) V
\end{equation}
where $Q = eW_{q1}$, $K = x_1W_{k1}$, $V = x_1W_{v1}$ are the query, key, and values matrices of the attention operation, and $W_{q1}$ , $W_{k1}$, $W_{v1}$ are the weight of the trainable linear projection layers.
% 考虑到生成的图片mask所引导的编辑部分需要和图二的某一具体事物高度相似，我们使用了linear layer来将feature from subject image $x_2$ 映射为 $f_2$ 而不是直接通过cross attn和 feature from subject image $x_2$与output from qformer 进行融合。

Considering that the the edited part in final image need to closely resemble a specific object in subject image $x_2$,  we utilized the linear layer and the decouple cross attention to integrate these features separately, rather than directly fusing them through cross-attention.
The subject feature is obtained by mapping the text embedding from $x_2$ through a linear layer. 
And then the features $f_1$ and $f_2$ go into the UNet block with decouple cross attention where the cross-attention layers. We add a new cross-attention layer for each cross-attention layer in the UNet block to for features interaction. Finally, given the query features $Z$, scene features $f_1$ and subject features $f_2$, the output of cross-attention $Z'$ can be defined by the following equation: 
\begin{equation}
    Z' = \text{Softmax}\left(\frac{Q K_1^\top}{\sqrt{d}}\right) V_1 + \lambda\text{Softmax}\left(\frac{Q K_2^\top}{\sqrt{d}}\right) V_2
\end{equation}
where $Q = ZW_{q2}$, $K_1 = f_1W_{k2}$, $V_1 = f_1W_{v2}$ and $K_2 = f_2W_{k2}$, $V_2 = f_2W_{v2}$ are the query, key, and values matrices of the attention operation respectively, and $W_{q2}$ , $W_{k2}$, $W_{v2}$ are the weight matrices of the trainable linear projection layers. $\lambda$ is weight factor, and we set $\lambda = 0$ when editing single image.

\subsection{Training to Understand Free-Shape Mask}
\label{sec:training_understand}

For easy annotation and efficient training, we modified the classical segmentation dataset COCO~\cite{coco} and GRefCOCO\textcolor{red}~\cite{gres} for free-shape mask understanding training process. Specifically, we first adapt the original mask label $m_o$ into free-shape mask $m=W(m_o)$ by a simple random walking algorithm $W$, and model are trained to predict the original mask while output the label of the predict class $p$. For example, the output can be ``The masks means `chair', \textless $img0\dots N$\textgreater'', \textless $img0\dots N$\textgreater~represent the generate image with original mask $m_o$ (see Figure~\ref{fig:mask_training}). More technical details will be discussed in the supplementary material.

\begin{figure}[ht]
    \centering
    \includegraphics[height=0.6\linewidth]{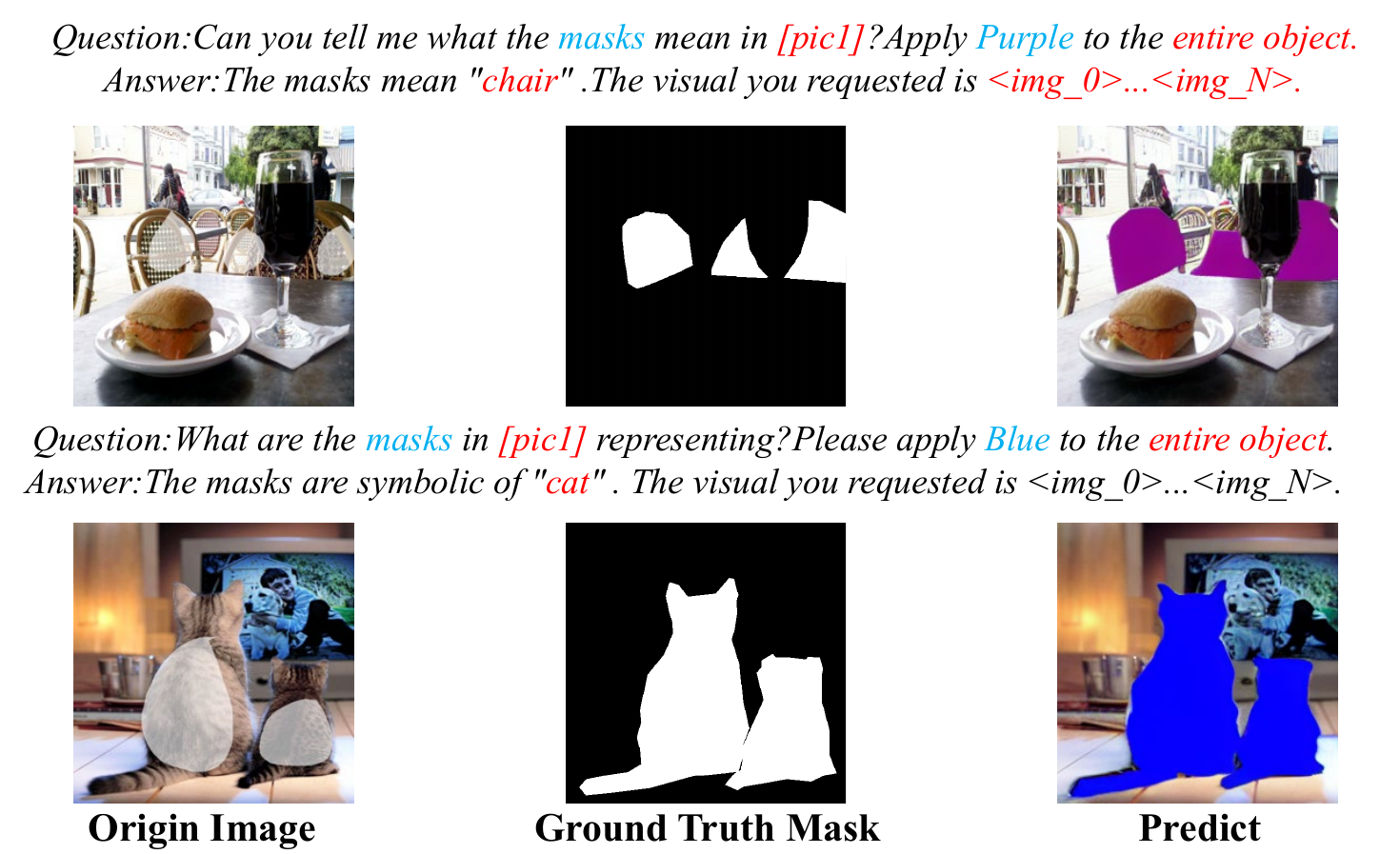}
    \caption{\small We train model to understand free-shape mask by having it predict the full mask from a given free-form mask. The picture shows two examples of this process.}
    \label{fig:mask_training}
\end{figure}

\subsection{FSMI-Edit Benchmark}
\label{sec:fsmi}

\begin{figure}
    \centering
    \includegraphics[width=0.9\linewidth]{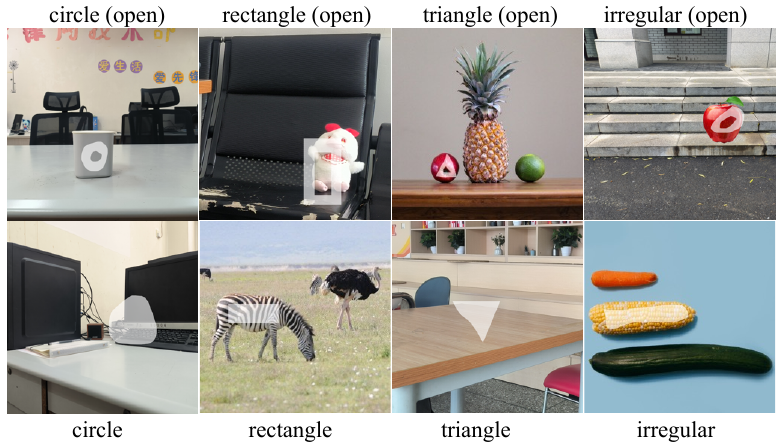}
    \caption{\small Example of the different types of the free-shape mask from FSMI-Edit benchmark.}
    \label{fig:fsmi_shows}
\end{figure}

% 修改开始：加强了关于该基准新颖性的声明。
To address the lack of standardized evaluation for this task, we construct the \textbf{F}ree-\textbf{S}hape \textbf{M}ask \textbf{I}nstruction \textbf{Edit} (FSMI-Edit) benchmark. To our knowledge, it is the first benchmark specifically designed to evaluate image editing under the realistic condition of imprecise, free-shape masks.
% 中文翻译如下：
% 为了解决该任务缺乏标准化评估的问题，我们构建了“自由形状掩模指令编辑”（FSIM-Edit）基准。据我们所知，这是首个专门设计用于在不精确、自由形状掩模这一真实条件下评估图像编辑的基准。
% 修改结束
We focus on daily indoor (e.g. Backpack, Kitchen) and outdoor (e.g. Animals, Ground) scenarios, for subject image we also consider two types of both simple (only subject provided) and complex (subject with noisy background) subject image. FSMI-Edit consists of 80 and 125 single images and multiple images editing picture respectively. We carefully design the mask type so that it can mimic human input preference, thoroughly evaluating the performance of free-shape mask reference image editing. Example of the free-shape mask from FSMI-Edit is shown at Figure~\ref{fig:fsmi_shows}. The free-shape mask consists of 8 type total: \textit{circle, circle(open hole), rectangle, rectangle (open hole), triangle, triangle (open hole), irregular, irregular (open hole) mask}, with different orientation. We hope more researchers will pay attention to the free-shape mask guided image editing from these perspectives, thereby fostering the user-friendly multi-modal instruction image editing methods. 

% \subsection{FSMI-Edit Benchmark}
% \label{sec:fsmi}

% We construct the \textbf{F}ree-\textbf{S}hape \textbf{M}ask \textbf{I}nstruction Edit Benchmark (FSMI-Edit) for evaluating image editing method in free-shape mask understanding. We focus on daily indoor (e.g. Backpack, Kitchen) and outdoor (e.g. Animals, Ground) scenarios, for subject image we also consider two types of both simple (only subject provided) and complex (subject with noisy background) subject image. FSMI-Edit consists of 80 and 125 single images and multiple images editing picture respectively.

% We carefully design the mask type so that it can mimic human input preference, thoroughly evaluating the performance of free-shape mask reference image editing. Example of the free-shape mask from FSMI-Edit is shown at Figure~\ref{fig:fsmi_shows}. The free-shape mask consists of 8 type total: \textit{circle, circle(open hole), rectangle, rectangle (open hole), triangle, triangle (open hole), irregular, irregular (open hole) mask}, with different orientation. We hope more researchers will pay attention to the free-shape mask guided image editing from these perspectives, thereby fostering the user-friendly multi-modal instruction image editing methods.

% \begin{figure}
%     \centering
%     \includegraphics[width=0.9\linewidth]{images/benchmark_shows.pdf}
%     \caption{\small Example of the different types of the free-shape mask from FSMI-Edit benchmark.}
%     \label{fig:fsmi_shows}
% \end{figure}

\section{Experiments}
\begin{table*}[t]
\centering
\caption{\small Comparison in multiple image editing domain, we compare the results in full mask and free-shape mask setting.}
\begin{adjustbox}{max width=0.9 \textwidth}
\begin{tabular}{lcccccccc}
\toprule
\multirow{3}{*}{\textbf{Method}} & \multicolumn{4}{c}{\textbf{Foreground}} & \multicolumn{4}{c}{\textbf{Whole Image}} \\
\cmidrule(lr){2-5} \cmidrule(lr){6-9}
 &\textbf{PNSR $\uparrow$} & \textbf{CLIP-T $\uparrow$} & \textbf{CLIP-I $\uparrow$} & \textbf{DINOv2 $\uparrow$} & \textbf{PNSR $\uparrow$} &\textbf{LPIPS $\downarrow$}& \textbf{CLIP-I $\uparrow$} & \textbf{DINOv2 $\uparrow$}  \\
 \midrule
\textit{\textbf{Full Mask}} &  &  &  &  &  & \\
AnyDoor & 13.1316 & 22.0478 & \underline{0.8959} & 0.2485 & \underline{19.2274} & \underline{0.1055} & 0.2340 & \underline{0.9292} \\
Largen & 12.5130 & 20.8105 & 0.8332 & 0.2468 & 18.6747 & 0.1166  & 0.2331 & 0.8886 \\
InstructAny2Pix & 8.6129 & 20.9438 & 0.7557 & 0.2424 & 10.3434 & 0.5231 & 0.1723 & 0.7100 \\
\cellcolor{gray!30}{\textbf{FlexEdit}} & \cellcolor{gray!30}{\underline{13.1645}} & \cellcolor{gray!30}{\underline{24.1754}} & \cellcolor{gray!30}{0.8279} & \cellcolor{gray!30}{\underline{0.2982}} & \cellcolor{gray!30}{18.0522} & \cellcolor{gray!30}{0.1433} & \cellcolor{gray!30}{\underline{0.2402}} & \cellcolor{gray!30}{0.8596} \\
\midrule
\textit{\textbf{Free-Shape Mask}} &  &  &  &  &  &  &  & \\
AnyDoor & 11.8483 & 19.5245 & 0.7711 & 0.2451 & 18.4441 & \textbf{0.1348} & 0.2280 & \textbf{0.8292} \\
Largen & 12.1133 & 19.1737 & 0.7627 & 0.2386 & \textbf{18.6563} & 0.1373 & 0.2266 & 0.8271 \\
InstructAny2Pix & 9.6056 & 20.5750 & 0.7582 & 0.2411 & 11.4099 & 0.4720 & 0.1775 & 0.7151 \\
\cellcolor{gray!30}{\textbf{FlexEdit}} & \cellcolor{gray!30}{\textbf{12.5648}} & \cellcolor{gray!30}{\textbf{22.4813}} & \cellcolor{gray!30}{\textbf{0.7795}} & \cellcolor{gray!30}{\textbf{0.2908}} & \cellcolor{gray!30}{17.9068} & \cellcolor{gray!30}{0.1585} & \cellcolor{gray!30}{\textbf{0.2353}} & \cellcolor{gray!30}{0.8262} \\
\bottomrule
\end{tabular}
\end{adjustbox}
\label{tab:multi_images}
\end{table*}

We compare our FlexEdit with SOTA models in both the single and multiple image editing domains. For a fair comparison, we fine-tune the models on the same dataset we use. For models that cannot be fine-tuned, we also report their performance under standard mask settings to assess their ability to understand the content when mask provided. 
\subsection{Experiments Setting}
\label{sec:ex_setting}
\subsubsection{Training Datasets}
\label{sec:training_dataset}

\begin{figure*}[ht!]
    \centering
    \includegraphics[width=1.0\linewidth]{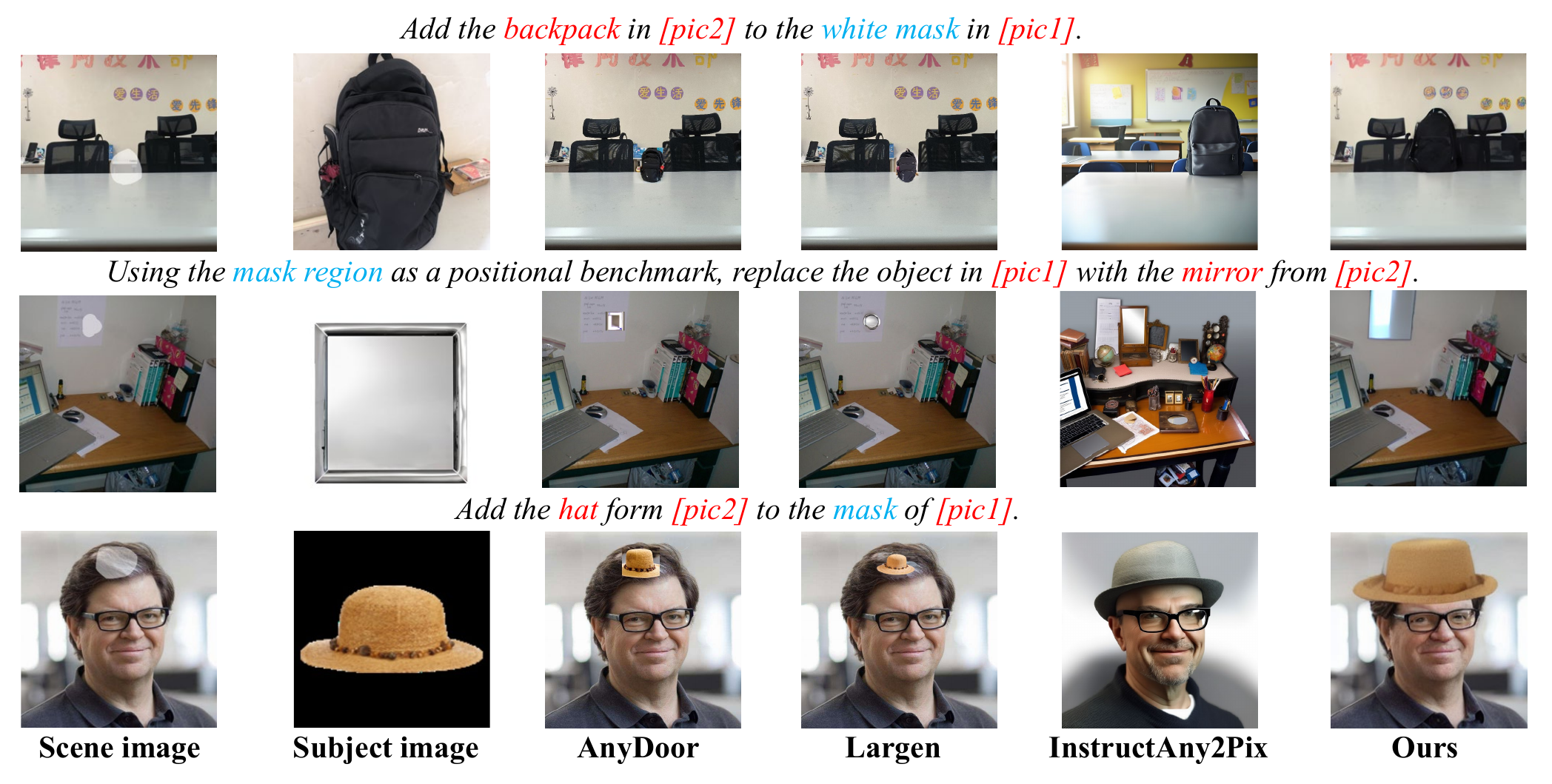}
    \caption{\small Qualitative results for free-shape mask guided image editing in multiple image setting.}
    \label{fig:model_comparison}
\end{figure*}

% \subsection{Training Datasets}
% We follow the setting of SmartEdit~\cite{smartedit} and include the training data of 4 categories: (1) segmentation datasets, which modified from COCO~\cite{coco} and GRefCOCO~\cite{gres};(2) Image editing datasets, which modified from MagicBrush~\cite{magicbrush},ReasonEdit~\cite{smartedit}, STRAT~\cite{STRAT};(3) In-Context visual question answering (VQA) dataset, which is the Mimic-it~\cite{Mimic} for it has multi-images input;(4) Our self-constructed image editing dataset, which we have collected a total of 311 pairs of images, sourced from real-life scenes and generated using other AI methods. Notably, 
% 但考虑用户在现实场景标注mask的往往比较随意这一情况，我们并不会把真实的mask直接作为输入，为了模仿人涂画我们在真实mask上使用随机游走的方式生成raw free-shape mask，加上人工筛选和二次标注，确保尽可能地模拟现实生活场景中的用户输入。而在分割数据集中，我们的利用策略是将原有的mask经过修改，以保证其更解决用户的真实输入，再通过模型来生成与GT 中形状一致的mask。而同时提供VQA dataset 能令大模型在微调的同时保持原有的分析和推理能力。更多的技术细节请查看附录。
% \textcolor{red}{

% 修改开始：为训练数据增加了百分比明细。
We include training data from four categories, with a large-scale, mixed composition: (1) 65% from image editing datasets (e.g., MagicBrush~\cite{magicbrush}, ReasonEdit~\cite{smartedit}, and STRAT~\cite{STRAT}); (2) 15% from segmentation datasets (e.g., COCO and GRefCOCO); (3) 10% from an in-context visual question answering (VQA) dataset (Mimic-it~\cite{Mimic}), chosen for its capability of handling multi-image inputs; and (4) 10% from our self-constructed image editing dataset, which includes a total of 390 pairs of images with complex subject images (e.g. ``Add the backpack in [pic2] to the white mask in [pic1]'' see Figure~\ref{fig:model_comparison}). 
% 中文翻译如下：
% 我们纳入了四类训练数据，其构成是一个大规模的混合数据集：（1）65% 来自图像编辑数据集（例如，MagicBrush~\cite{magicbrush}, ReasonEdit~\cite{smartedit}, 和 STRAT~\cite{STRAT}）；（2）15% 来自图像分割数据集（例如，COCO 和 GRefCOCO）；（3）10% 来自上下文视觉问答（VQA）数据集（Mimic-it~\cite{Mimic}），选择它是因为它能处理多图像输入；（4）10% 来自我们自建的图像编辑数据集，其中包含总共390对带有复杂主题图像的图片（例如，“将[pic2]中的背包添加到[pic1]的白色掩模中”，见图~\ref{fig:model_comparison}）。
% 修改结束
To address the imprecision of user-drawn masks in real-world applications, we avoid using the actual masks directly as inputs. Instead, we simulate user input by generating free shape masks using random walks on the real masks. These generated masks are then reviewed and re-annotated by three volunteers to ensure they closely mimic real-life user inputs. For the segmentation datasets, the model's objective is to regenerate masks based on the free-form input to match the ground truth shapes. Training on the VQA dataset ensures that the model retains its analytical and reasoning capabilities during fine-tuning. Additional technical details can be found in the supplementary materials. 

% We include training data from four categories: (1) segmentation datasets, which are modified from COCO and GRefCOCO; (2) image editing datasets, which are adapted from MagicBrush~\cite{magicbrush}, ReasonEdit~\cite{smartedit}, and STRAT~\cite{STRAT}; (3) in-context visual question answering (VQA) datasets, specifically Mimic-it~\cite{Mimic}, due to its capability of handling multi-image inputs; (4) our self-constructed image editing dataset, which includes a total of 390 pairs of images with complex subject image that should be learned from understand the prompt (e.g. ``Add the backpack in [pic2] to the white mask in [pic1]'' see Figure~\ref{fig:model_comparison}).

% To address the imprecision of user-drawn masks in real-world applications, we avoid using the actual masks directly as inputs. Instead, we simulate user input by generating free shape masks using random walks on the real masks. These generated masks are then reviewed and re-annotated by three volunteers to ensure they closely mimic real-life user inputs. For the segmentation datasets, the model's objective is to regenerate masks based on the free-form input to match the ground truth shapes. Training on the VQA dataset ensures that the model retains its analytical and reasoning capabilities during fine-tuning. Additional technical details can be found in the supplementary materials.

\subsubsection{Metrics}
% \subsection{Metrics}
%我们希望模型在编辑的foreground尽量在遵循指令的同时聚焦于mask区域，并且不要过度影响整个图片。但由于用户给出的mask是较为随意的，所以我们不能直接采用用户的mask来作为前景和背景的区分，而是将GT中的实际编辑部分作为划分的mask。我们选择了CLIP-T，CLIP-I，DINO v2这三个指标来评价多图编辑的前景部分。CLIP-T用于评价前景部分与GT text label的契合度，GT text label is annotated manually. CLIP-I和DINO v2用于评价生成图片和GT中的前景部分的图像相似度。同时选择PNSR，CLIP-I，DINO v2这三个指标来评价多图编辑整个图片与GT的相似度。而在单图编辑的指标中，遵循SmartEdit，我们选择了PSNR，SSIM，LPIPS用于背景的相似度的评估，而使用CLIP-T来评价编辑图像的前景部分和GT text label的契合度.
\label{sec:metrics}
The final objective for the model is to focus understand the free-shape mask area during editing, adhering to the instructions while minimizing changes to the overall image. To assess the ability in the mask understanding, we modified the mask-evaluate setting. Unlike the setting of aforementioned methods that use the mask area in evaluating the generate quality, given that free-shape masks' imprecise nature, we use the real edited regions from the GT as the evaluate masks instead of the input one. In thoroughly assessing the quality of image editing, we split the metric into the single and multiple image evaluation:

% \begin{itemize}
\textbf{Multiple Image:} We include PNSR, CLIP-T, CLIP-I~\cite{clip}, and DINOv2~\cite{dinov2} for the metric of foreground quality assessing, where CLIP-T measures how performance of the edited foreground aligns with the text label. PNSR, CLIP-I and DINOv2 assess the visual similarity between the generated images and the GT foreground images. LPIPS was introduced in the whole image evaluation while CLIP-T is excluded since CLIP-T label can not be annotated in the whole image setting.

\textbf{Single Image:} For single-image editing, in line with SmartEdit~\cite{smartedit}, we use PSNR, SSIM, LPIPS to evaluate background similarity, and CLIP-T to assess the alignment of the results' foreground with the text label.
% \end{itemize}

\subsection{Comparisons with SOTA models}
\label{sec:comparsion}
\subsubsection{Baselines}
We compared the diffusion only method AnyDoor and Largen, as well as the VLLM plus diffusion method instructAny2Pix, with our method in multiple image editing. Here, AnyDoor and Largen are the methods designed for zero-shot subject-driven image editing task, generating appropriate objects from the subject image within the masked area of the scene image. InstructAny2Pix leverages a VLLM to combine multiple user images and text instructions for editing the corresponding images.

% 这种方法能在scene image的maks area内生成合适的object from subject image. 而instructAny2Pix能够通过视觉大模型来结合用户的多张图片和text instruction来编辑的对应图片。
For single image editing, we compare FlexEdit with InstructPix2Pix, InstructAny2Pix, and SmartEdit.
%InstructPix2Pix and SmartEdit 通过human language instruction来对图片进行编辑， InstructionPix2Pix使用clip结构来对人类指令和图片内容进行理解同时SmartEdit使用视觉大语言模型来对指令和图片内容理解。
InstructPix2Pix and SmartEdit edit images based on human language instructions. InstructPix2Pix uses a CLIP structure to understand human instructions and image content, while SmartEdit employs a large visual language model to interpret human instructions and image content.

% We compare our FlexEdit with state-of-the-art models in the single and multiple image editing domains. For a fair comparison, we fine-tune the models on the same dataset we use. For models that cannot be fine-tuned, we also report the scores under the normal mask settings, evaluating they can understand the content if the mask provided.

% 多图编辑的实验结果如表一所示，单图编辑的实验结果如表二所示。

\subsubsection{Multiple Image Editing}
We first provide a qualitative comparison between FlexEdit with other method with add/replace operation and simple/complex subject images in Figure~\ref{fig:model_comparison}. The compared model either fail to understand the free-shape mask (AnyDoor, Largen), or unintentionally change the background image (InstructAny2Pix).

\textbf{Foreground Quality Comparison:} Under the normal mask setting, the AnyDoor and Largen shows the best score in CLIP-I, indicating the exhibit of the highest similarity to the foreground GT image, wheras InstructAny2Pix is less than ours. DINOv2 shows the same trend as CLIP-I, except ours FlexEdit achieved 0.2982, marginly higher than others. For the CLIP-T scores which represented the similarity between the generated foreground and the text label, however, FlexEdit achieved 24.1754, achieving the SOTA results. In the free mask setting, where user-provided masks are more arbitrary and abstract, methods like AnyDoor and Largen, which generate images only within the mask, fail to meet user expectations, resulting in a significant drop in metrics. InstructAny2Pix's performance do not decrease and perform stable in this setting. In this setting, our model have a sightly performance drop, while achieve the state-of-the-art results among all the metrics, showing our strong ability in understanding the free-shape mask. The detail results can be found at Table~\ref{tab:multi_images} (Foreground Image).

\textbf{Whole Image Quality Comparison:} The PNSR, CLIP-I, DINOv2 is close in the whole image setting for AnyDoor, Largen and FlexEdit. Note that InstructAny2Pix perform marginly bad at this setting, indicating that it will modify the unintentional background image, this is a long-stand problem in an end-to-end diffusion generation framework. However, our FlexEdit while edit the foreground object well, keeping the whole image stable, demonstrating our model's strong image understanding and image edting abilities. More detail results can be found at Table~\ref{tab:multi_images} (Whole Image).

% 在多图编辑领域，我们比较了AnyDoor 和Largen这两个仅基于diffusion的方法和instructAny2Pix将LLM和diffusion结合的方法。可以看到，normal mask settings的情况下，Foreground 下的CLIP(I)与DINOv2指标反映了AnyDoor和Largen方法编辑后的图片与前景图像相似度最高，CLIP(T)指标反映了这些图片与文字描述也保持了较高的一致性。
% 同时Whole Image下的三个指标（PNSR，CLIP(I),DINO v2）说明了编辑后的图像图像与GT整体图像的一致。而Instruct Any2Pix的表现均较差，由该方法编辑后的图片能够保证一定的前景图像的文字相似度和图像相似度，但是它对整体图像的扰动过大，变更了用户需求中不需要编辑的图像内容部分。而我们的方法编辑得到的图片在得到最高的文字相似度的同时保持了相对较高的图像相似度，同时保证了整体图像的一致性。
% 在free mask setting下，用户的提供的mask就会相对随意和抽象，这个时候anydoor和largen这类方法就会限制于他只在mask内生成图片的局限性，无法生成符合用户期望的图片，导致前景和全图指标的大幅度下降。而因为Instruct Any2Pix在free mask的场景下的整体指标与在normal mask下相比较为稳定。而我们的方法在free mask的场景下前景指标仍然维持了一个较好的水平，能够结合用户给出的mask，图片信息和具体指令，来对图片进行精确的编辑。同时whole image指标反映出我们的编辑方法没有过度影响用户所不希望修改的地方。fig 4的多图部分说明了这一点。

\subsubsection{Single Image Editing}
\begin{figure}
    \centering
    \includegraphics[width=1.0\linewidth]{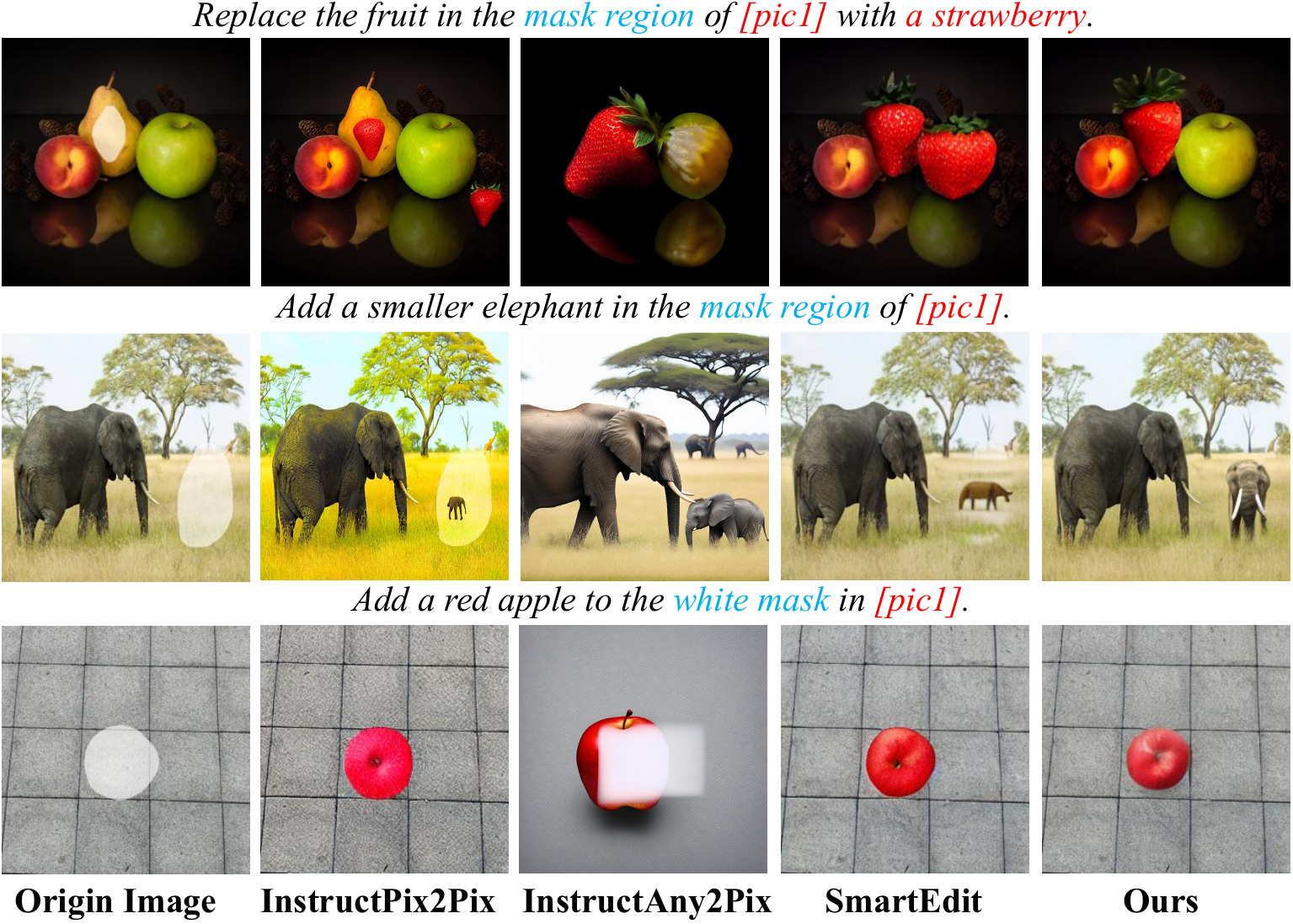}
    \caption{\small Qualitative results comparison on single image edit.}
    \label{fig:model_comparison_single}
\end{figure}

For a fair comparison, we fine-tune InstructPix2Pix and SmartEdit under the same setting as ours\footnotemark[1].
\footnotetext[1]{InstructAny2Pix has not released the training code at the time this paper submitted.} The qualitative comparison is shown in Figure~\ref{fig:model_comparison_single}, InstructAny2Pix can understand the free-shape mask, while will change the background image unintentionally. SmartEdit works better than InstructPix2Pix for it can understand the mask occasionally, instructionPix2Pix however, shows bad performance among the three methods.

Results shows that InstructPix2Pix performs bad even after fine-tuning, we argue that these can come from the problem that CLIP simple encoder can not fully understand the free-shape mask. InstrucAny2Pix still suffers from it can not maintain the whole image during the editing process. FlexEdit in the other hand, understand the free-shape mask and keep the rest of the image still, this can come from our MEA module for its ability in information extraction. SmartEdit shares the similar result as ours, with a sightly weakness that can come from our single-multiple image mixed training strategy. Detail results can be found at Table~\ref{tab:single_images}.
More comparison can be found in the supplementary material.

% 而在单图编辑领域中，我们比较了InstructPix2Pix,SmartEdit和Instruct Any2Pix这三种方法，其中InstructPix2Pix和SmartEdit都在我们的相同的单图数据集上微调过。可以看到Instruct Pix2Pix在所有的指标上都较低，简单的clip结构无法让他难以做到理解用户给出的free shape mask，并在编辑的同时不扰动图片的整体内容。而Instruct Any2Pix还是对整体图片的扰动过大，编辑了用户不想修改的部分。SmartEdit能够结合用户给出的mask和instruction对具体的图像内容进行编辑，但在总体指标上与我们的指标相比依然稍逊一筹。这可能是由于训练场景缺失更多复杂的多图场景导致的。
\begin{table}[ht]
\centering
\caption{\small Comparison in single image editing domain.}
\begin{adjustbox}{max width=1.2\textwidth}
\begin{tabular}{lcccc}
\toprule
\textbf{Method}& \textbf{PSNR $\uparrow$} & \textbf{SSIM $\uparrow$} & \textbf{LPIPS $\downarrow$} & \textbf{CLIP-T $\uparrow$} \\
\midrule
InstructPix2Pix & 19.9414 & \textbf{0.8016} & 0.1043 & 20.7010\\
InstructAny2Pix & 15.5149 & 0.5919 & 0.2751 & 20.3758\\
SmartEdit & 21.9535 & 0.7620 & 0.0781 & 20.4910 \\
\cellcolor{gray!30}{\textbf{FlexEdit}} & \cellcolor{gray!30}{\textbf{22.8081}} & \cellcolor{gray!30}{0.7712} & \cellcolor{gray!30}{\textbf{0.0727}} & \cellcolor{gray!30}{\textbf{20.7081}} \\
\bottomrule
\end{tabular}
\end{adjustbox}
\label{tab:single_images}
\end{table}

\begin{table*}[h]
\centering
\caption{\small Ablation study in multiple image editing. We evaluate the `CA', standing for cross attention in MEA, and `DCA', standing for decouple cross attention.}
\begin{adjustbox}{max width=1.0\textwidth}
\begin{tabular}{llcccccccc}
\toprule
\multirow{3}{*}{\textbf{CA}} & \multirow{3}{*}{\textbf{DCA}} & \multicolumn{4}{c}{\textbf{Foreground}} & \multicolumn{4}{c}{\textbf{Whole Image}} \\
\cmidrule(lr){3-6} \cmidrule(lr){7-10}
 & & \textbf{PNSR $\uparrow$} & \textbf{CLIP-T $\uparrow$} & \textbf{CLIP-I $\uparrow$} & \textbf{DINOv2 $\uparrow$} & \textbf{PNSR $\uparrow$} &\textbf{LPIPS $\downarrow$}& \textbf{CLIP-I $\uparrow$} & \textbf{DINOv2 $\uparrow$}  \\
 \midrule
\checkmark & \checkmark & \textbf{12.5794} & \textbf{22.4391} & \textbf{0.7796} & \textbf{0.2890} & \textbf{17.9068} & \textbf{0.1585} & \textbf{0.2353} & \textbf{0.8262} \\
- & \checkmark & 12.1836 & 21.6230 &  0.7451 & 0.2846  &17.5058 & 0.1806 & 0.2378 & 0.8140\\
\checkmark & - & 11.8367 & 21.3692 & 0.7561 & 0.2799 & 17.0976 & 0.1813 & 0.2341 & 0.8124\\
\bottomrule
\end{tabular}
\end{adjustbox}
\label{tab:multi_ablation}
\end{table*}

\subsection{Ablation study on MEA}
\label{ablation}
% 修改开始：澄清了消融研究的重点和基本原理。
To validate our contribution, the ablation study was intentionally focused on the MEA's key components: the Cross Attention (CA) and Decouple Cross Attention (DCA) modules. This follows the standard experimental practice of isolating variables to clearly demonstrate the effectiveness of a new component. We designed two variant structures and compared their performance metrics in both single and multiple image editing to validate the effectiveness of each module.
% 中文翻译如下：
% 为了验证我们的贡献，消融研究有意地聚焦于MEA的关键组件：交叉注意力（CA）和解耦交叉注意力（DCA）模块。这遵循了标准的实验惯例，即通过隔离变量来清晰地展示新组件的有效性。我们设计了两种变体结构，并在单图和多图编辑中比较了它们的性能指标，以验证每个模块的有效性。
% 修改结束
First, we remove the Cross Attention (CA) layer and directly used the output of the QFormer as the scene feature input into the following decouple cross attention, without the scene picture interaction. We also change the Decouple Cross Attention (DCA) structure with a general cross attention and examine the MEA effectiveness in image fusion. As shown in Table~\ref{tab:multi_ablation}, in multiple image editing, the CA and DCA can help boost the model performance, both seperately and jointly. For single image editing, as shown in Table~\ref{tab:single_ablation}, though the MEA module does not work in this mode, it can learn during the joint training process, and CA, DCA involving shows a better result for the single image performance. 

\begin{table}[!h]
\centering
\caption{\small Ablation study in single image editing, same setting as experiment in multiple image editing.}
\begin{adjustbox}{max width=1.0\textwidth}
\begin{tabular}{llcccc}
\toprule
\textbf{CA} & \textbf{DCA} & \textbf{PSNR $\uparrow$} & \textbf{SSIM $\uparrow$} & \textbf{LPIPS $\downarrow$} & \textbf{CLIP-T $\uparrow$} \\
\midrule
\checkmark & \checkmark & \textbf{22.8081} & \textbf{0.7712} & \textbf{0.0727} & \textbf{20.7081} \\
- & \checkmark & 21.8579 & 0.7546 & 0.0931 & 19.8630 \\
\checkmark & - & 20.4195 & 0.7230 & 0.1186 & 19.8566 \\
\bottomrule
\end{tabular}
\end{adjustbox}
\label{tab:single_ablation}
\end{table}
\FloatBarrier

% 修改开始：增加了关于局限性和失败案例的新章节。
\subsection{Limitations and Failure Case Analysis}
Despite the strong performance of FlexEdit, our work has some limitations and potential areas for future improvement.

\textbf{Choice of VLLM Backbone.} We chose LLaVA to ensure a controlled and fair comparison with SOTA methods like SmartEdit, allowing us to isolate the benefits of our proposed contributions. While this provides a clear comparison, future work could explore performance improvements with alternative or more recent VLLM backbones, such as Qwen-VL\cite{qwen}.

\textbf{Failure Cases.} We have identified three primary modes of failure:
\begin{itemize}
    \item \textbf{VLLM Grounding Failure:} When a free-shape mask is highly ambiguous (e.g., covering two distinct objects simultaneously), the VLLM may misinterpret the user's intent, leading to an incorrect edit.
    \item \textbf{Semantic Misinterpretation:} In some cases, the model may generate an object that is semantically related to the instruction but is contextually incorrect (e.g., generating a tiger when a cat was requested).
    \item \textbf{Background Artifacts:} Minor, unintended changes to the background style or texture can occasionally occur. This is a known challenge in end-to-end diffusion generation frameworks that can affect the overall image quality.
\end{itemize}

\section{Conclusion}

In this paper, we introduce FlexEdit, an end-to-end method based on Visual Language Models (VLLM) for free shape mask guided image editing. Our model addresses the challenge of imprecise and irregular mask shapes drawn by humans by incorporating free shape masks during the training process. To achieve effective image fusion, we propose a Mask Enhanced Adapter (MEA), which enhanced the mask embedding from the VLLM with image data for seamless blending. To support the image editing community, we also develop FSMI-Edit, a new evaluation dataset specifically designed for free shape mask scenarios. Our method achieves state-of-the-art results on the FSMI-Edit benchmark, outperforming other models trained under similar conditions, whether in free shape mask or full-mask settings.

\newpage

\bibliographystyle{plainnat}
\bibliography{reference}

\end{document}